\documentclass{article}

\usepackage[preprint,nonatbib]{neurips_data_2023}
\usepackage{times}
\usepackage{epsfig}
\usepackage{graphicx}
\usepackage{amsmath}
\usepackage{amssymb}
\usepackage{caption}
\usepackage{graphicx}
\usepackage{float} 

\usepackage{subcaption}
\usepackage{rotating}
\usepackage{diagbox}
\usepackage{graphicx}
\usepackage{tabularx}
\usepackage{float}
\usepackage{alphalph}

\usepackage[utf8]{inputenc} 
\usepackage[T1]{fontenc}    
\usepackage{hyperref}       
\usepackage{url}            
\usepackage{booktabs}       
\usepackage{amsfonts}       
\usepackage{nicefrac}       
\usepackage{microtype}      
\usepackage{xcolor}         
\usepackage{float}

\usepackage{setspace}
\title{Reliable Evaluation of Adversarial Transferability}

\author{
Wenqian Yu$^{1}$\thanks{Equal Contribution} \;\;
Jindong Gu$^{2}$\footnotemark[1] \;\;
Zhijiang Li$^{1}$\thanks{Corresponding Author} \; \;
Philip Torr$^{2}$ \; \\
$^{1}$Wuhan University,
$^{2}$University of Oxford\\
\texttt{jindong.gu@eng.ox.ac.uk, lizhijiang@whu.edu.cn}\\}

\begin{document}

\maketitle

\begin{abstract}
Adversarial examples (AEs) with small adversarial perturbations can mislead deep neural networks (DNNs) into wrong predictions. The AEs created on one DNN can also fool another DNN. Over the last few years, the transferability of AEs has garnered significant attention as it is a crucial property for facilitating black-box attacks. Many approaches have been proposed to improve adversarial transferability. However, they are mainly verified across different convolutional neural network (CNN) architectures, which is not a reliable evaluation since all CNNs share some similar architectural biases. In this work, we re-evaluate 12 representative transferability-enhancing attack methods where we test on 18 popular models from 4 types of neural networks. Our reevaluation revealed that the adversarial transferability is often overestimated, and there is no single AE that can be transferred to all popular models. The transferability rank of previous attacking methods changes when under our comprehensive evaluation. Based on our analysis, we propose a reliable benchmark including three evaluation protocols. Adversarial transferability on our new benchmark is extremely low, which further confirms the overestimation of adversarial transferability. We release our benchmark at \url{https://adv-trans-eval.github.io} to facilitate future research, which includes code, model checkpoints, and evaluation protocols.
\end{abstract}

\section{Introduction}
Adversarial examples (AEs) are inputs with imperceptible perturbations to cause the model to make incorrect predictions~\cite{2013Intriguing,2015Explaining}. AEs created to attack one model can also fool other models, which is known as adversarial transferability. The transferability has received great attention recently since it can implement attacks without access to target models. In addition, studying adversarial transferability can also help to better understand the difference between models.

In recent years, the transferability has been intensively studied between CNNs, such as VGGs~\cite{simonyan2014very}, ResNets~\cite{he2016deep}, Inceptions~\cite{szegedy2015going} and Densenets~\cite{huang2017densely}, as a means of evaluating the transferability of attack methods. As many improvements have been proposed~\cite{TI,DIM,zou2020improving, wu2021improving,wu2020skip,LinBP,ILA,IR,3d2023transfer}, it has become easier to transfer AEs between CNNs. However, recent studies~\cite{naseer2021improving,mahmood2021robustness,wei2022towards} indicate that the transferability between Vision Transformers (ViTs)~\cite{dosovitskiy2020image} and CNNs is poor, highlighting the need to investigate black-box transferability between various types of models. Such investigations would enable a more reliable evaluation of adversarial transferability, thereby offering valuable insights.

To this end, we conducted an extensive evaluation of the transferability of AEs across 18 different models, encompassing four popular model types: CNNs~\cite{lecun1998gradient}, ViTs~\cite{dosovitskiy2020image}, Spiking Neural Networks (SNNs)~\cite{tavanaei2019deep}, and Dynamic Neural Networks (DyNNs)~\cite{han2021dynamic,wu2016deep,wang2020glance}. Our experiments included 12 attack methods. By studying the impact of model architectures, attack methods, and attack settings on transferability, we obtained the following findings: (1) Transferability between the same type of models is usually higher than that between different types of models. The model hyper-parameters can have a big impact on the transferability effectiveness, e.g., patch size and inference steps. (2) The effectiveness rank of transferable attacks becomes different when different source models and target models are evaluated, and even can be inverse. Many recently proposed improvements in adversarial transferability are questioned by our comprehensive evaluation. (3) The more transferable attack methods do not necessarily perform better when a new perturbation range setting is used, which urges a comprehensive evaluation in future research.

To provide a more reliable and comprehensive way for evaluating adversarial transferability, we propose a new benchmark that includes three testing protocols. Protocol-1 is a hard-ensemble model which combines all tested target models by averaging their logits, creating a new 'confidence score' for the ensemble model. Protocol-2 counts the number of models that an AE can transfer to. In protocol-3, an AE is considered to transfer successfully only when it can transfer to all the target models. Adversarial transferability on our new benchmark is lower compared with that obtained on CNNs which confirms that the transferability is overestimated in previous works. 

Our contributions can be briefly summarized as follows: 1) We conducted a study on transferability from CNNs to three other popular models: ViTs, SNNs, and DyNNs. We tested 18 target models using 12 different transferability-enhancing attack methods. 2) We proposed a more reliable and comprehensive benchmark for evaluating adversarial transferability. The transferability on our new benchmark confirms the overestimation of adversarial transferability. 3) Based on our novel findings and protocols, we analyzed the existing attack methods. The effectiveness of these attack methods is different from the results evaluated by previous work.

\section{Related Work}
\textbf{Transfer-based Attack Methods.} Extensive studies have been conducted to improve the transferability of AEs across CNN~\cite{lecun1998gradient} models and many attack methods have been proposed from the perspective of data, models, loss design, and optimization methods. The work~\cite{TI,DIM,zou2020improving, wu2021improving} aims to overcome the problem of over-fitting to the source model with transformed inputs. In addition,\cite{wu2020skip,LinBP,ghost-network} fine-tune the structures of source models when generating AEs. Moreover, several novel designs of loss functions~\cite{ILA,IR,inkawhich2020perturbing,zhang2022improving} have been proposed to enhance the transferability of AEs. These designs approach the problem from different perspectives rather than relying solely on the cross-entropy loss from ground truth. Apart from models and loss, the optimization methods have also been improved, like the momentum methods~\cite{MI}, Nesterov Accelerated Gradient~\cite{Lin2020Nesterov} and variance tuning~\cite{wang2021enhancing}.

\textbf{Evaluation of Adversarial Transferability.} Evaluating adversarial transferability is crucial to assessing the robustness of machine learning models, the similarity of different models, and the effectiveness of attack methods. One common approach for evaluating adversarial transferability is generating AEs using a source model and then testing their effectiveness against a target model. The success rate of these attacks can be used as a metric for evaluating the transferability of the AEs. Most studies use classic CNNs~\cite{simonyan2014very,he2016deep,szegedy2015going,huang2017densely} as the source and target models. In recent years, several studies have explored transferability across models other than CNNs. The work~\cite{naseer2021improving,mahmood2021robustness,wei2022towards} observed transferability across ViTs~\cite{dosovitskiy2020image} and CNNs and introduced novel attack methods specific to ViTs' structures. Xu~\cite{xu2022securing} analyzed transferability across SNNs, ViTs and CNNs and developed a new white-box attack. The works~\cite{wu2022towards,gu2022vision,pinto2022impartial,benz2021adversarial} compared the adversarial robustness of ViT to CNNs. The robustness difference between CNNs and Capsule Networks is also studied~\cite{gu2020improving,gu2021capsule}. However, despite a small number of studies on transferability across models other than CNNs, transferability between other novel architectures remains underexplored. This limitation makes it difficult for the research community to reliably evaluate adversarial transferability across all popular architectures.

\section{Preliminary Knowledge}
\subsection{Transferable Attack Methods}

We first introduce the basic adversarial attack methods and then present the transferability-enhancing attacks from 4 perspectives, namely, data-based, model-based, loss-based, and optimization-based.

\textbf{Adversarial Attack.} Let ${f}$ be a K-class classification model, the input of which is an image $X \in {\mathbb{R}^{H \times W \times C}}$ and each image has a one-hot label ${y^{true}} \in {\mathbb{R}^{1 \times K}}$. The clean image $X^{clean}$ is classified as ${f}({{X^{clean}}})$. Adversarial attack aims to fool the model into making incorrect predictions by adding imperceptible perturbations and the constrained optimization problem can be written as $\mathop {\arg \max }_{{X^{adv}}} \mathcal{L}({{f}({{X^{adv}}})},{y^{true}}), \; s.t. \;{\left\| {{X^{adv}} - {X^{clean}}} \right\|_\infty } \leqslant \varepsilon$
where $\mathcal{L}( \cdot )$ is the loss function which is often the cross entropy loss. $\varepsilon $ is the allowed perturbation range and ${\left\|  \cdot  \right\|_\infty }$ is the ${L_\infty }$ norm of distance between the clean and adversarial image. 

FGSM~\cite{2015Explaining} generates AEs in the direction of the loss gradient and performs single-step update as ${X^{adv}} = {X^{clean}} + \varepsilon  \cdot sign({\nabla _X}\mathcal{L}({{f}({{X^{clean}}})},\;{y^{true}}))$ where ${\nabla _X}\mathcal{L}$ is the gradient of the loss function to the input and $sign( \cdot )$ is the sign function.
PGD~\cite{madry2017towards} is the multi-step variant of FGSM, i.e., $X_{t + 1}^{adv} = \phi(X_t^{adv} + \alpha  \cdot sign({\nabla _X}\mathcal{L}({f}(X_t^{adv}),\;{y^{true}})))$ where $X_t^{adv}$ is the AEs generated on t-th iteration and $\alpha$ is the step size of each iteration. $\phi(\cdot)$ is the projection operation to satisfy the constraint ${\left\| {{X^{adv}} - {X^{clean}}} \right\|_\infty } \leqslant \varepsilon$. 

\textbf{Data-based Transferable Attacks.}  The methods~\cite{TI,DIM,zou2020improving,wu2021improving,Byun_2022_CVPR,Wang_2021_ICCV,li2020regional,huang2022transferable,Integrated_gradients} apply transformations to the images before they are input into the classifier at each iteration: $X_{t + 1}^{adv} = \phi(X_t^{adv} + \alpha  \cdot sign({\nabla _X}\mathcal{L}(f(T(X_t^{adv})),{y^{true}})))$, where $T( \cdot )$ is the transform function. The diversity input attack (DI Attack) and translation-invariant attack (TI Attack) are based on data augmentation. TI~\cite{TI} utilizes the translation-invariant property of CNNs~\cite{lecun1998gradient} and optimizes a perturbation over an ensemble of translated images, which is essentially a gradient-blurring method. DI~\cite{DIM} applies random transformations to the inputs, including random resizing and padding with a certain probability. To tackle the limitation of simple image transformations, more advanced transform methods have been proposed in recent years, such as RDI~\cite{zou2020improving}, ODI~\cite{Byun_2022_CVPR}, Admix~\cite{Wang_2021_ICCV} and ATTA~\cite{wu2021improving}. ODI breaks the limitation of 2D transformation and projects images on 3D objects; Admix abandons transformation on a single image, admixing inputs with images from other categories; ATTA adopts the idea of GAN~\cite{goodfellow2020generative} and uses CNN to simulate adversarial transformations.

\textbf{Model-based Transferable Attacks.} The architectures of source models are slightly modified in order to prevent AEs from over-fitting to them~\cite{wu2020skip,LinBP,ghost-network,salzmann2021learning,rozsa2017lots,inkawhich2020transferable,waseda2023closer,benz2021batch,liu2017delving,gubri2022lgv,li2023making,zhu2021rethinking}. The skip gradient method (SGM Attack)~\cite{wu2020skip} improves the proportion of skip-connection gradients, while the linear back-propagation method (LinBP Attack)~\cite{LinBP} skips the nonlinear layer in the back-propagation process. There exist other attacks related to the structure of models, like IAA~\cite{zhu2021rethinking} and ghost network~\cite{ghost-network}. IAA smooths activation layers with softplus function; ghost network ensembles models with different parameters of dropout and skip-connection. 

\textbf{Loss-based Transferable Attacks.} Several new losses, different from traditional ones, have been proposed to boost adversarial transferability. Novel loss functions enrich the features involved and alleviate the problem of AEs over-fitting to the particular architecture and feature representations of a source model. The feature maps from intermediate layers are utilized in some attack methods~\cite{ILA,IR,inkawhich2020perturbing,zhou2018transferable,inkawhich2019feature,naseer2018task,ganeshan2019fda,wang2021feature,fang2022learning,zhao2021success,zhang2022investigating,xiao2021improving,li2020towards}:  the intermediate level attack (ILA)~\cite{ILA} fine-tunes the AEs by increasing the perturbation on a pre-specified layer of the source model. The interaction-reduced attack (IR)~\cite{IR} uses interaction loss, which penalizes interaction inside adversarial perturbations. The new loss function can boost transferability due to the negative correlation between transferability and interaction. Apart from adding new items to the loss function~\cite{fang2022learning,zhao2021success}, novel normalization or regularization methods are also effective, like new normalized cross-entropy loss~\cite{zhang2022investigating}, using generative models~\cite{xiao2021improving} or metric learning~\cite{li2020towards} as regularization.

\textbf{Optimization-based Transferable Attacks.} Methods based on advanced optimization algorithms have also progressed to prevent over-fitting of AEs to the source models. The seminal momentum iterative attack (MI)~\cite{MI} adds the momentum term  when calculating gradients to accelerate gradient descent and escape local optima. The variance tuning momentum iterative attack (VMI)~\cite{wang2021enhancing} considers the gradient variances at each iteration based on the MI method. The Nesterov iterative attack (NI)~\cite{Lin2020Nesterov} intergrates Nesterov Accelerated Gradient into the iterative process to improve convergence. In addition to these, more and more various novel optimization methods have emerged~\cite{xu2022adversarially, zou2022making, xiong2022stochastic, qin2022boosting}: AI-FGTM~\cite{zou2022making} replaces sign with Tanh function; SVRE~\cite{xiong2022stochastic} reduces the gradient variance of the ensemble models during optimization; RAP~\cite{qin2022boosting} seeks the flat minimum of loss value instead of the sharp one by injecting the reverse adversarial perturbation for each step of the optimization procedure.

\subsection{Models of Different Neural Architectures}
\label{models}
\textbf{Convolutional Neural Networks.}
CNNs~\cite{lecun1998gradient} utilize convolutional layers to extract spatial features from input images, leveraging weight sharing. Among popular architectures such as VGG16~\cite{simonyan2014very}, ResNet50~\cite{he2016deep}, DenseNet121~\cite{huang2017densely}, and InceptionV3~\cite{szegedy2015going}, each model has its own distinctive characteristics, with InceptionV3 employing parallel convolutional filters to capture features at different scales, while ResNets and DenseNets utilize skip connections to address the vanishing gradient and over-fitting problem, enabling deeper learning and improved performance.

\textbf{Vision Transformers.}
ViTs~\cite{dosovitskiy2020image}, unlike CNNs, represent an image as a list of patches, utilizing self-attention mechanisms to extract global features rather than being restricted by local features like convolution. Popular variants of ViTs are Data-efficient Image Transformers (DeiTs)~\cite{touvron2021training}, and Vision Transformers with shifted windows (Swin Transformers)~\cite{liu2021swin}. DeiTs achieve data efficiency through knowledge distillation, while Swin Transformers use a hierarchical network architecture with shifted windows to capture context information efficiently for large-scale image processing tasks.

\textbf{Spiking Neural Networks.}
SNNs~\cite{tavanaei2019deep} have attracted significant attention in recent years due to their energy-efficient and biologically-inspired computing. The concept of time is incorporated into SNNs. Neurons in the SNN do not transmit information continuously at each propagation cycle, but rather fire discrete spikes only when the accumulated stimulus reaches the threshold. Learning mechanisms in SNNs are unique due to the non-differentiable nature of spiking neurons and can be categorized into unsupervised learning~\cite{diehl2015unsupervised}, supervised learning using surrogate gradients~\cite{wang2020supervised,taherkhani2018supervised,wang2014online}, and conversion-based training~\cite{rueckauer2017conversion,li2022converting,li2021free}. The conversion-based training method involves mapping pretrained CNN parameters to an SNN and fine-tuning the weights to improve performance.

\textbf{Dynamic Neural Networks.}
Different from traditional static neural networks with a fixed topology and number of parameters, DyNNs~\cite{han2021dynamic,wu2016deep} can adapt their structure and behavior in response to the complexity of the problem, resulting in efficiency and less computational cost. The DyNN tested in our experiment is Glance and Focus Network(GFNet)~\cite{wang2020glance}. The whole process of GFNet is sequential, glance and focus being two stages. Glance processes the down-scaled images and those with discriminate features can be classified with high confidence at this stage. If the prediction is not confident enough, the framework moves on to the focus stage, where it processes progressively smaller, class-discriminative regions of the full-resolution image. The focus stage allows for adaptive termination based on prediction confidence.

\section{Benchmark and Analysis of Transferable Attack Methods}

\subsection{Evaluation Settings}

\noindent\textbf{Evaluation Dataset.}
We randomly selected 5000 images from the ILSVRC 2012 validation set~\cite{deng2009imagenet} that were correctly classified by all the tested source models. The images are resized according to the input sizes of tested models. The input size of all tested models is 224 except InceptionV3~\cite{szegedy2015going}, whose input size is 299.

\noindent\textbf{Evaluation Model.}
18 models are included as target models. Among them, 7 representative ones are selected as source models. The notations of models are as follows: For CNNs~\cite{simonyan2014very,he2016deep,szegedy2015going,huang2017densely}, Res50 is composed of a model name and the number of layers; For ViTs~\cite{dosovitskiy2020image,touvron2021training}, ViT-S/16 means the 'Small' variant with 16×16 input patch size. Swin transformer~\cite{liu2021swin} is notated with one more parameter window size, e.g. Swin-B/4/7; For SNNs~\cite{li2022converting}, SNN-Res34/L is a representative name where SNN are converted from Res34 and 'Light' (L) means only the bias is calibrated. GFNets~\cite{wang2020glance} are also based on CNNs and the parameters are patch size (96 and 128) and step (5 in default). The evaluation mode was set to 1, following~\cite{wang2020glance}. See Supplement A for more details.

\noindent\textbf{Attack Methods.}
12 attack methods are included. PGD~\cite{madry2017towards}, TI~\cite{TI}, DI~\cite{DIM}, MI~\cite{MI}, NI~\cite{Lin2020Nesterov}, and VMI~\cite{wang2021enhancing} are applied to all source models; LinBP~\cite{LinBP} can only be used on Res50~\cite{he2016deep}; SGM~\cite{wu2020skip} is applied to two source models (Res50 and Dense121~\cite{huang2017densely}); ILA~\cite{ILA} is also applied to two source models (Dense121 and IncepeionV3~\cite{szegedy2015going}); IR~\cite{IR} is used on four CNN models; Self-Ensemble (SE)~\cite{naseer2022on} and Pay No Attention (PNA)~\cite{Wei_Chen_Goldblum_Wu_Goldstein_Jiang_2022} are ViT-specific attacks. For PGD attack, We set $\varepsilon  = 16/255$, $step size = 2/255$ for ${L_\infty }$ attack with 40 iterations. Detail of each attack can be found in Supplement B.

\noindent\textbf{Evaluation Metric.} The standard metric \textbf{Fooling Rate (FR)} is applied to evaluate the transferability. We denote $P$ as the number of adversarial images that successfully fool a source model. Among them, the number of images that are also able to fool a target model is $Q$. The Fooling Rate is then defined as FR = $\frac{Q}{P}$. The higher the FR is, the more transferable the adversarial images are.

\subsection{Evaluation Results}

In this section, we describe evaluation results and present some findings from our analysis. 

\begin{figure}[t]
    \centering
    \includegraphics[width=0.96\linewidth]{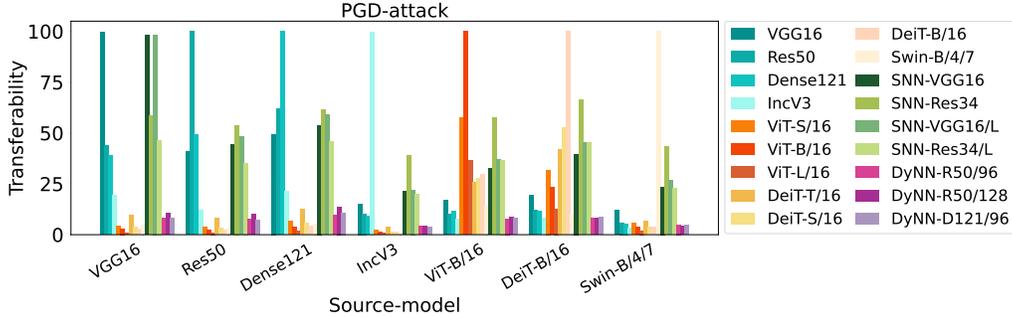}
    \caption{Transferability of AEs created by PGD\cite{madry2017towards} attack on different models. It shows transferability (FR scores) from 7 source models to 18 target models. The horizontal axis represents source models, while the vertical axis represents Fooling rate. Transferability within the same type of models is relatively high and SNNs~\cite{tavanaei2019deep} are the weakest models. It is difficult to attack ViTs~\cite{dosovitskiy2020image} with AEs generated on CNNs.}
    \label{fig1}  \vspace{-0.5cm}   
\end{figure}

\subsubsection{Impact of Source and Target Models on Adversarial Transferability}
As shown in Fig.~\ref{fig1}, transferability from CNNs~\cite{lecun1998gradient} to other CNNs and SNNs~\cite{li2022converting} is higher than that to DyNNs~\cite{wang2020glance}, while the transferability to transformers~\cite{dosovitskiy2020image,touvron2021training,liu2021swin} is the lowest. This is not surprising, given that SNNs are converted based on CNNs, and DyNNs can be seen as dynamic combinations of CNN blocks. The transformer, using self-attention mechanism instead of convolution operations, is the model with the greatest difference from CNNs, hence its lower transferability.

\textbf{Analysis on CNN.}
Fig.~\ref{fig1} and the first 4 rows of Tab.~\ref{trans-res50} show that the transferability between Res50~\cite{he2016deep} and Dense121~\cite{huang2017densely} is high, which is expected since both models are equipped with skip connections. Among the four CNN models tested, IncV3~\cite{szegedy2015going} is less vulnerable to AEs created on other models (low fooling rates on IncV3), as shown in Fig.~\ref{fig1}. We conjecture that this is caused by its unique design of factorizing convolutions. Additionally, the transferability from IncV3 to VGG16~\cite{simonyan2014very} is higher than that to other CNNs with skip connections, which indicates that IncV3 and VGG16 learn similar functions that differ from the residual learning-based models. Similar conclusions can also be drawn under other attack methods, as shown in Tab.~\ref{trans-res50}.

\begin{table}[!ht]
    \scriptsize
    \centering
    \caption{Transferability with Res50~\cite{he2016deep} as the source model. Each row shows FRs on a target model, while each column represents an attack method. The largest FR value of each row is marked in bold. SGM~\cite{wu2020skip} and LinBP~\cite{LinBP} are the most effective on Res50.  TI~\cite{TI} and DI~\cite{DIM} get high FR to DyNNs~\cite{wang2020glance}. * marks the attack success rate on the source model where IR~\cite{IR} gets the lowest score.}
    \begin{tabularx}{{0.993\linewidth}}{c|ccccccccc}
    \hline
        \diagbox{{Target}}{{Attacks}} & {PGD~\cite{madry2017towards}} & {TI~\cite{TI}} & {DI~\cite{DIM}} & {SGM~\cite{wu2020skip}} & {LinBP~\cite{LinBP}} & {IR~\cite{IR}} & {MI~\cite{MI}} & {NI~\cite{Lin2020Nesterov}} & {VMI~\cite{wang2021enhancing}} \\ \hline
        {VGG16} & 40.86 & 31.16 & 76.30 & 85.40 & \textbf{91.50} & 66.34 & 61.96 & 61.92 & 56.88 \\ 
        Res50* & 100.00* & 99.98* & 100.00* & 99.98* & 100.00* & 92.84* & 99.98* & 99.98* & 99.06* \\ 
        {Dense121} & 49.04 & 41.16 & 85.40 & 84.62 & \textbf{92.88} & 77.18 & 67.12 & 67.00 & 63.00 \\ 
        {IncV3} & 12.18 & 14.18 & 43.54 & 42.50 & \textbf{55.60} & 34.40 & 27.58 & 27.80 & 25.02 \\ \hline
        {ViT-S/16} & 3.82 & 6.32 & 12.06 & \textbf{21.58} & 15.16 & 12.66 & 10.08 & 9.64 & 8.50 \\ 
        {ViT-B/16} & 2.34 & 3.56 & 7.12 & \textbf{13.40} & 9.50 & 7.82 & 6.16 & 5.88 & 5.34 \\ 
        {ViT-L/16} & 1.02 & 1.94 & 2.88 & \textbf{4.78} & 2.74 & 3.30 & 2.84 & 3.04 & 2.70 \\ 
        {Deit-T/16} & 8.16 & 10.70 & 18.86 & \textbf{30.34} & 23.34 & 18.96 & 16.92 & 16.68 & 15.90 \\ 
        {Deit-S/16} & 3.24 & 4.76 & 9.80 & \textbf{16.82} & 10.28 & 9.54 & 9.10 & 9.22 & 8.20 \\ 
        {Deit-B/16} & 2.38 & 2.86 & 7.06 & \textbf{12.42} & 7.38 & 7.78 & 6.42 & 6.32 & 5.68 \\ 
        {Swin-B/4/7} & 2.86 & 1.74 & 8.78 & \textbf{14.58 }& 11.30 & 9.58 & 5.52 & 5.96 & 4.78 \\ \hline
        {SNN-VGG16} & 44.14 & 33.10 & 68.50 & 73.76 & \textbf{84.32} & 60.28 & 68.60 & 67.72 & 63.22 \\ 
        {SNN-Res34} & 53.80 & 68.70 & 66.50 & 63.90 & \textbf{79.26} & 63.20 & 76.00 & 76.66 & 72.90 \\ 
        {SNN-VGG16/L} & 48.22 & 41.48 & 72.90 & 77.52 & \textbf{88.44} & 65.40 & 73.12 & 73.22 & 68.50 \\ 
        {SNN-Res34/L} & 35.26 & 49.08 & 58.00 & 59.74 & \textbf{79.74} & 51.86 & 63.56 & 65.00 & 59.60 \\ \hline
        {DyNN-R50/96} & 7.65 & \textbf{23.68} & 16.92 & 19.20 & 17.51 & 12.56 & 18.03 & 18.25 & 17.32 \\ 
        {DyNN-R50/128} & 10.11 & 27.61 & \textbf{35.56} & 34.00 & 34.45 & 23.68 & 25.43 & 25.34 & 24.10 \\ 
        {DyNN-D121/96} & 7.32 & \textbf{26.62} & 17.90 & 17.57 & 18.03 & 13.40 & 16.79 & 16.88 & 17.26 \\ \hline
    \end{tabularx}
    \label{trans-res50} \vspace{-0.4cm}
\end{table}

\begin{table}[t]
    \scriptsize
    \centering
    \caption{Transferability with ViT-B/16~\cite{dosovitskiy2020image} as the source model. Similar to Tab.~\ref{trans-res50}, the largest FR and scores in the white-box setting are marked. DI obtains the highest scores for most target models. MI and NI have the highest FR when SNNs~\cite{li2022converting} are used as targets.}
    \setlength\tabcolsep{12pt}
    \begin{tabularx}{{0.957\linewidth}}{c|cccccc}
    \hline
        \diagbox{{Target}}{{Attacks}}& {PGD~\cite{madry2017towards}} & {TI~\cite{TI}} & {DI~\cite{DIM}} & {MI~\cite{MI}} & {NI~\cite{Lin2020Nesterov}} & {VMI~\cite{wang2021enhancing}} \\ \hline
        {VGG16} & 16.96 & 14.60 & 38.82 & 39.34 & \textbf{40.54} & 28.20 \\
        {Res50} & 10.20 & 10.44 & \textbf{36.66} & 21.58 & 22.16 & 14.78 \\ 
        {Dense121} & 11.40 & 11.60 & \textbf{38.80} & 21.84 & 21.28 & 14.32 \\ 
        {IncV3} & 7.82 & 8.14 & \textbf{31.80} & 15.04 & 15.54 & 9.48 \\ 
        \hline
        {ViT-S/16} & 57.58 & 39.78 & \textbf{87.76} & 62.74 & 62.22 & 47.20 \\ 
        {ViT-B/16*} & 100.00* & 100.00* & 100.00* & 100.00* & 100.00* & 95.14* \\ 
        {ViT-L/16} & 36.36 & 28.46 & \textbf{78.60} & 49.60 & 49.58 & 32.96 \\ 
        {Deit-T/16} & 25.58 & 16.94 & \textbf{53.06} & 35.80 & 36.34 & 25.18 \\ 
        {Deit-S/16} & 27.64 & 15.98 & \textbf{60.34} & 38.34 & 38.52 & 26.62 \\ 
        {Deit-B/16} & 29.54 & 15.58 & \textbf{63.78} & 38.42 & 39.34 & 24.90 \\ 
        {Swin-B/4/7} & 14.02 & 6.88 & \textbf{53.36} & 18.70 & 18.86 & 10.40 \\ 
        \hline
        {SNN-VGG16} & 32.50 & 24.04 & 50.68 & \textbf{62.26} & 61.30 & 50.18 \\ 
        {SNN-Res34} & 57.32 & 62.44 & 68.68 & 80.18 & \textbf{80.42} & 73.98 \\ 
        {SNN-VGG16/L} & 36.90 & 30.32 & 56.78 & \textbf{68.94} & 68.86 & 56.14 \\ 
        {SNN-Res34/L} & 36.72 & 38.16 & 55.76 & 66.96 & \textbf{67.68} & 53.02 \\ 
        \hline
        {DyNN-R50/96} & 7.91 & 15.46 & \textbf{19.82} & 16.96 & 17.00 & 13.78 \\ 
        {DyNN-R50/128} & 8.47 & 13.59 & \textbf{28.30} & 17.43 & 17.94 & 12.88 \\ 
        {DyNN-D121/96} & 8.14 & 17.49 & \textbf{21.42} & 15.91 & 16.18 & 13.12 \\ \hline
    \end{tabularx}
    \label{trans-vit} \vspace{-0.4cm}
\end{table}

\textbf{Analysis on ViT.}
The transferability from ViTs~\cite{dosovitskiy2020image,touvron2021training,liu2021swin} to both CNNs and DyNNs~\cite{wang2020glance} is low in general since they are based on different building blocks. However, the transferability to SNNs~\cite{li2022converting} is surprisingly high, which is caused by the intrinsic vulnerability of SNNs. Please see the analysis on SNNs. Among the three variants of transformers, the transferability between ViTs~\cite{dosovitskiy2020image} and DeiTs~\cite{touvron2021training} is higher than that to Swin transformers~\cite{liu2021swin} (row 5-11 of Tab.~\ref{trans-vit}). This observation is intuitive since ViTs and DeiTs have similar architecture, while Swin transformers adopt a totally different hierarchy architecture and shifted-window-based attention. 

Besides the architectures of transformers, the impact of model size and input patch size to transferability is also studied. The first observation is that transferability decreases as the models become larger. Compared with model size, the influence of patch size is greater especially when transferring between transformers. The transferability between transformers with the same patch size is extremely high. When transferred to CNNs, AEs on ViTs with different patch sizes do not show any pattern. All the supporting scores are in Supplement C.

\textbf{Analysis on DyNN.}
Tab.~\ref{trans-res50} also suggests that the transferability from CNNs~\cite{simonyan2014very,he2016deep,szegedy2015going,huang2017densely} to DyNNs~\cite{wang2020glance} is higher for DyNNs with larger patch sizes, as shown in the 16th and 17th rows of the table. This could be due to the fact that CNNs process the entire image, while DyNNs only select patches of images for classification in later stages. As the patch size increases, the patch becomes closer to the original image, resulting in DyNNs with larger patch sizes being more similar to CNNs.

\begin{figure}[t]
	\centering
	\begin{subfigure}{0.317\linewidth}
		\centering
		\includegraphics[width=0.8\linewidth]{figure/res50-DyNN_Res50_P96.pdf}
		\caption{DyNN-R50/96}
		\label{gfnet-res50-1}
	\end{subfigure}
	\begin{subfigure}{0.2765\linewidth}
		\centering
		\includegraphics[width=0.8\linewidth]{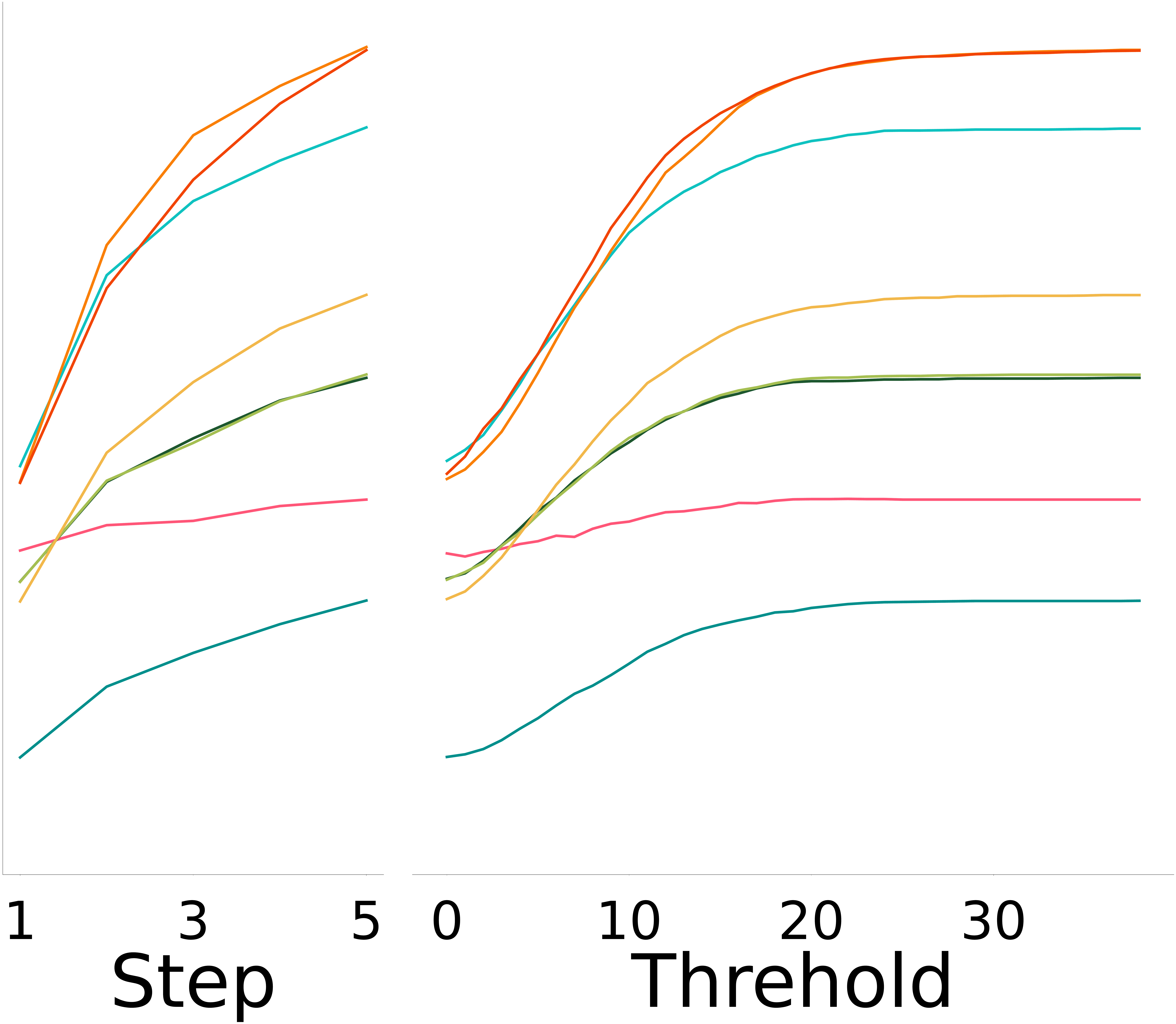}
		\caption{DyNN-R50/128}
		\label{gfnet-res50-2}
	\end{subfigure}
        \begin{subfigure}{0.3766\linewidth}
		\centering
		\includegraphics[width=0.8\linewidth]{figure/res50-DyNN_Dense121_P96.pdf}
		\caption{DyNN-D121/96}
		\label{gfnet-res50-3}
	\end{subfigure}
        \caption{Transferability from CNN-Res50~\cite{he2016deep} to three GFNets~\cite{wang2020glance} with different hyperparameters and attack methods. It comprises 3 sub-figures, each describing the transferability to one GFNet variant. X-axis is composed of two parts. The first part corresponds to the number step 1 to 5, and the second one to thresholds. Almost all attacks generate more transferable AEs as the number of steps or the threshold becomes larger.}
	\label{gfnet-res50} \vspace{-0.3cm}
\end{figure}

The impact of DyNNs' hyperparameters is also studied, such as the number of steps, threshold, and patch size. As shown in Fig.~\ref{gfnet-res50}, the transferability becomes higher with the increase of inference steps and threshold under most attacks. The possible reason is that the input scale of the first step is different from that of adversarial noises and the following steps. We choose the results of threshold-0 for evaluation, where most attacks attain the lowest transferability except for TI~\cite{TI}. Hence, TI gets the best performance when transferring to GFNets~\cite{wang2020glance}. Comparing Fig.~\ref{gfnet-res50-1} and ~\ref{gfnet-res50-2}, we can find that the FR increases when the patch size of the GFNet becomes larger.

\begin{table}[t]
\centering
\scriptsize
\caption{Transferability rank of attack methods with different models. It shows that the rank can change when different source or target models are evaluated.}
\label{rank-change-table}
\setlength\tabcolsep{2.8pt}
\begin{tabularx}{{0.468\linewidth}}{c|c|c}
\hline 
{Source Models} & {Target Models} & {Rank of Transferability}  \\ \hline
{CNNs} & CNNs \& ViTs & IR~\cite{IR} > MI~\cite{MI} \& NI~\cite{Lin2020Nesterov}  \\ 
{CNNs} & SNNs \& DyNNs & IR~\cite{IR} < MI~\cite{MI} \& NI~\cite{Lin2020Nesterov}  \\\hline
{Res50} & CNNs \& SNNs & LinBP~\cite{LinBP} > SGM~\cite{wu2020skip}  \\
{Res50} & ViTs & LinBP~\cite{LinBP} < SGM~\cite{wu2020skip}  \\ \hline
\end{tabularx} \hspace{0.1cm}
\begin{tabularx}{{0.425\linewidth}}{c|c|c}
\hline 
{Source Models} & {Target Models} & {Rank of Transferability}  \\ \hline
{Dense121} & CNNs \& SNNs & ILA~\cite{ILA} > SGM~\cite{wu2020skip}  \\
{Dense121} & ViTs \& DyNNs & ILA~\cite{ILA} < SGM~\cite{wu2020skip}  \\ \hline
{Res50} & ViTs & TI~\cite{TI} > PGD~\cite{madry2017towards} \\
{ViTs} & ViTs & TI~\cite{TI} < PGD~\cite{madry2017towards} \\ \hline
\end{tabularx} \vspace{-0.5cm}
\end{table}

\textbf{Analysis on SNN.}
It is shown in Fig.~\ref{fig1} that when CNNs or transformers~\cite{dosovitskiy2020image,touvron2021training,liu2021swin} are used as the source model, the transferability of AEs to SNNs~\cite{li2022converting} is relatively high. It indicates that the robustness of SNN models to adversarial perturbation is poor. Spiking neurons might be responsible for such results because their outputs can change a lot despite tiny input perturbations. Meanwhile, AEs created by MI~\cite{MI} and NI~\cite{Lin2020Nesterov} attacks show high transferability to SNNs, regardless of the source models, which is supported by 12th-15th rows in Tab.~\ref{trans-res50} and Tab.~\ref{trans-vit}.

\subsubsection{Impact of Attack Methods on Adversarial Transferability}
In addition to the architecture of the models, the effectiveness of adversarial attacks is also impacted by the attack methods employed. When transferring from CNNs~\cite{simonyan2014very,he2016deep,szegedy2015going,huang2017densely} to transformers~\cite{dosovitskiy2020image,touvron2021training,liu2021swin}, the AEs generated with data-based attack methods such as TI~\cite{TI} and DI~\cite{DIM} tend to exhibit lower transferability, while those generated with model-based methods like SGM~\cite{wu2020skip} and LinBP~\cite{LinBP} tend to have higher transferability. The possible reason is that data-based attacks prevent over-fitting by averaging or blurring gradients and in contrast, model-based methods obtain more efficient gradients from the perspective of model architecture directly.

However, the TI~\cite{TI} and DI~\cite{DIM} attack methods have high transferability from CNNs to DyNNs~\cite{wang2020glance} (shown in the last three rows of Tab.~\ref{trans-res50}) while TI has poor transferability to other types of target models. Overall, as illustrated in the 6th column of Tab.~\ref{trans-res50}, IR~\cite{IR} yields relatively higher transferability for most target models, but the white-box attack success rates are lower than other methods. It can be understood as a trade-off between black-box and white-box attack success rates. 

\textbf{Does the rank of attack methods hold the same across all evaluation models?} No.
In terms of adversarial transferability, the rank of attack methods will change when AEs are transferred to different models, as shown in Tab.~\ref{rank-change-table}. Surprisingly, there exist many pairs of attack methods whose rank will be reversed when the target or source models change. E.g., from CNNs to CNNs and ViTs (common setting), AEs generated with IR~\cite{IR} are more transferable than those with MI~\cite{MI}. However, MI is more effective than IR when transferring from CNNs to SNNs and DyNNs.

\textbf{Do ViT-specific transferable attacks improve transferability to all models?} Yes. Two representative ViT-specific transferable attacks are tested in our experiments: self-ensemble (SE)~\cite{naseer2022on} and Pay No Attention (PNA)~\cite{Wei_Chen_Goldblum_Wu_Goldstein_Jiang_2022} attack. SE attack inputs class tokens from all blocks within ViT into the shared classifier and sums the cross-entropy losses up as the self-ensemble loss. PNA attack skips the gradients of attention during back-propagation to prevent over-fitting, which is similar to SGM~\cite{wu2020skip} attack for CNNs. As illustrated in Tab.~\ref{vit-attack}, the two attack methods do improve the transferability to CNNs. Specifically, when CNNs and DyNNs are target models(the first four rows and last three rows in Tab.~\ref{vit-attack}), except for VGG16~\cite{simonyan2014very}, PNA can obtain more transferable AEs compared with SE attack. However, when transferring to SNNs~\cite{li2022converting}, AEs generated with SE get higher transferability than those generated with PNA, which is inconsistent with the conclusion in~\cite{Wei_Chen_Goldblum_Wu_Goldstein_Jiang_2022}.

\begin{table}[t]
    \centering
    \scriptsize
    \caption{Transferability on ViT-B/16~\cite{dosovitskiy2020image} under ViT-specific attacks. Two improving attacks (SE~\cite{naseer2022on} and PNA~\cite{Wei_Chen_Goldblum_Wu_Goldstein_Jiang_2022}) based on basic attacks (PGD~\cite{madry2017towards}, TI~\cite{TI}, DI~\cite{DIM}, MI~\cite{MI} and NI~\cite{Lin2020Nesterov}) are shown. SE and PNA can improve transferability to all target models. * indicates white-box attack.}
    \setlength\tabcolsep{2.7pt}
    \begin{tabularx}{{0.955\linewidth}}{c|ccc|ccc|ccc|ccc|ccc}
    \hline
        \diagbox{{Target}}{{Attack}} & {PGD} & {+SE} & {+PNA} & {TI} & {+SE} & {+PNA} & {DI} & {+SE} & {+PNA} & {MI} & {+SE} & {+PNA} & {NI} & {+SE} & {+PNA} \\ \hline
        {VGG16} & 16.96 & \textbf{30.24} & 26.08 & 14.60 & 24.62 & \textbf{24.66} & 38.82 & \textbf{61.84} & 53.18 & 39.34 & \textbf{49.68} & 48.02 & 40.54 & \textbf{49.72} & 48.02 \\ 
        {Res50} & 10.20 & 14.94 & \textbf{16.72} & 10.44 & 14.34 & \textbf{18.54} & 36.66 & 47.38 & \textbf{51.08} & 21.58 & 25.86 & \textbf{28.40} & 22.16 & 25.92 & \textbf{29.22} \\
        {Dense121} & 11.40 & 16.30 & \textbf{17.26} & 11.60 & 16.96 & \textbf{21.26} & 38.80 & 52.68 & \textbf{53.30} & 21.84 & 26.16 & \textbf{28.30} & 21.28 & 26.84 & \textbf{29.12} \\ 
        {IncV3} & 7.82 & 10.88 & \textbf{11.48} & 8.14 & 11.04 & \textbf{13.72} & 31.80 & 45.84 & \textbf{46.22} & 15.04 & 17.86 & \textbf{19.42} & 15.54 & 17.90 & \textbf{19.00} \\ \hline
        {ViT-S/16} & 57.58 & \textbf{72.32} & 68.88 & 39.78 & 49.48 & \textbf{53.30} & 87.76 & \textbf{95.12} & 93.44 & 62.74 & \textbf{77.38} & 72.88 & 62.22 & \textbf{77.48} & 72.70 \\ 
        {ViT-B/16*} & 100* & 100* & 100* & 100* & 100* & 100* & 100* & 100* & 100* & 100* & 100* & 100* & 100* & 100* & 100* \\
        {ViT-L/16} & 36.36 & 32.02 & \textbf{37.78} & 28.46 & 26.04 & \textbf{34.62} & 78.60 & 77.76 & \textbf{81.58} & 49.60 & 43.60 & \textbf{50.10} & 49.58 & 43.84 & \textbf{50.28} \\ 
        {Deit-T/16} & 25.58 & \textbf{42.14} & 37.50 & 16.94 & 25.82 & \textbf{28.80} & 53.06 & \textbf{74.96} & 68.62 & 35.80 & \textbf{54.78} & 48.06 & 36.34 & \textbf{54.48} & 48.60 \\
        {Deit-S/16} & 27.64 & 35.82 & \textbf{37.82} & 15.98 & 18.46 & \textbf{26.78} & 60.34 & 74.14 & \textbf{74.36} & 38.34 & 48.82 & \textbf{49.24} & 38.52 & \textbf{48.90} & 48.62 \\
        {Deit-B/16} & 29.54 & 32.56 & \textbf{38.38} & 15.58 & 15.38 & \textbf{24.96} & 63.78 & 72.46 & \textbf{75.76} & 38.42 & 41.84 & \textbf{46.36} & 39.34 & 41.54 & \textbf{47.20} \\
        {Swin-B/4/7} & 14.02 & 9.58 & \textbf{16.22} & 6.88 & 5.06 & \textbf{8.70} & 53.36 & 49.90 & \textbf{58.12} & \textbf{18.70} & 12.66 & 18.26 & \textbf{18.86} & 13.14 & 18.02 \\ \hline
        {SNN-VGG16} & 32.50 & \textbf{56.38} & 46.04 & 24.04 & \textbf{40.98} & 36.18 & 50.68 & \textbf{77.50} & 66.24 & 62.26 & \textbf{76.26} & 69.68 & 61.30 & \textbf{76.34} & 69.48 \\
        {SNN-Res34} & 57.32 & \textbf{76.34} & 68.96 & 62.44 & \textbf{76.20} & 73.70 & 68.68 & \textbf{85.68} & 80.30 & 80.18 & \textbf{87.70} & 84.48 & 80.42 & \textbf{87.32} & 84.76 \\
        {SNN-VGG16/L} & 36.90 & \textbf{61.92} & 51.80 & 30.32 & \textbf{50.24} & 46.66 & 56.78 & \textbf{84.26} & 74.20 & 68.94 & \textbf{81.76} & 76.90 & 68.86 & \textbf{81.54} & 75.98 \\
        {SNN-Res34/L} & 36.72 & \textbf{61.18} & 53.78 & 38.16 & \textbf{60.08} & 58.74 & 55.76 & \textbf{85.20} & 74.86 & 66.96 & \textbf{80.82} & 77.16 & 67.68 & \textbf{80.74} & 77.08 \\ \hline
        {DyNN-R50/96} & 7.80 & 11.42 & \textbf{11.46} & 12.68 & 22.92 & \textbf{26.24} & 20.54 & 26.38 & \textbf{28.54} & 17.56 & 20.76 & \textbf{21.80} & 17.68 & 20.56 & \textbf{21.68} \\ 
        {DyNN-R50/128} & 8.54 & 12.46 & \textbf{13.56} & 12.78 & 20.84 & \textbf{25.58} & 29.06 & 39.56 & \textbf{41.38} & 18.32 & 22.62 & \textbf{24.36} & 18.64 & 22.88 & \textbf{24.36} \\
        {DyNN-D121/96} & 8.56 & 12.02 & \textbf{12.80} & 14.14 & 25.76 & \textbf{30.52} & 22.50 & 29.94 & \textbf{31.46} & 16.88 & 20.34 & \textbf{20.84} & 16.98 & 20.38 & \textbf{21.38} \\ \hline
    \end{tabularx}
    \label{vit-attack} \vspace{-0.4cm}
\end{table}

\subsubsection{Impact of Attack Settings on Adversarial Transferability}
\textbf{Targeted vs. Untarget.}
The results under targeted attack are shown in Supplement D. For each image, the target class is randomly sampled from 999 classes excluding the ground-truth class. The white-box attack success rate of targeted attack is similar to that of untarget attack. However, it is quite difficult for targeted AEs to transfer and FRs are nearly zero for most models and attack methods. DI attack on ViT-B/16~\cite{dosovitskiy2020image} and DeiT-B/16~\cite{touvron2021training} performs the best among ViT variants.

\textbf{Perturbation Range.}
For most attack methods, transferability gets higher with the increase of epsilon values, which is shown in Supplement E. However, IR~\cite{IR} attack is the exception: AEs generated with epsilon 16 are more transferable to CNNs than those generated with epsilon 32. It is also worth mentioning that the transferability rank also changes under evaluation with different epsilons.

\textbf{Perturbation Visibility.} We visualize adversarial perturbations and calculate their L2-norm. L2-norm scores on IncV3~\cite{szegedy2015going} is larger than those on other source models. When the source model is the same, the differences between L2-norm values under different attacks are relatively similar. Furthermore, there is little correlation between transferability and L2-norm, as shown in Supplement G. As shown in visualization in Supplement H, perturbations can be focused on objects in some cases.

\begin{table}[t]
    \scriptsize
    \centering
    \caption{Scores of three protocols in our new benchmark. Each column represents one source model and each row represents one attack method. There are three scores (protocol-1 to 3 from left to right) in each cell. The highest scores for each source model are marked in bold. DI~\cite{DIM} shows the highest scores in all three protocols when ViTs~\cite{dosovitskiy2020image,touvron2021training,liu2021swin} are used as source models.}
    \label{protocol}
    \setlength\tabcolsep{3.0pt}
    \begin{tabularx}{{1.0\linewidth}}{c|c|c|c|c|c|c|c}
    \hline
        {Model} & VGG16 & Res50 & Dense121 & IncV3 & ViT-B/16 & Deit-B/16 & Swin-B/4/7 \\ \hline
        PGD & 4.74/ 5.49/ 0.06 & 2.68/ 4.22/ 0.10 & 5.52/ 5.19/ 0.16 & 0.62/ 2.42/ 0.04 & 7.22/ 5.11/ 0.40 & 4.54/ 5.31/ 0.26 & 0.76/ 2.66/ 0.06 \\ 
        TI & 7.54/ 5.73/ 0.12 & 4.86/ 4.81/ 0.14 & 7.62/ 5.27/ 0.22 & 1.80/ 3.60/ 0.08 & 6.64/ 4.54/ 0.22 & 3.92/ 4.43/ 0.42 & 3.72/ 3.99/ 0.34 \\ 
        DI & 9.88/ 6.59/ 0.34 & 14.24/ 7.01/ 0.68 & \textbf{20.10}/ \textbf{7.82}/ 0.82 & 2.20/ 3.73/ 0.04 & \textbf{36.62}/ \textbf{9.40}/ \textbf{5.56} & \textbf{27.74}/ \textbf{9.49}/ \textbf{5.28} & \textbf{10.40}/ \textbf{5.45}/ \textbf{1.08} \\ 
        SGM & / & \textbf{21.62}/ 7.74/ \textbf{1.22} & 16.22/ 6.95/ \textbf{0.98} & / & / & / & / \\ 
        LinBP & / & 13.44/ \textbf{8.29}/ 0.52 & / & / & / & / & / \\ 
        ILA & / & / & 7.64/ 6.66/ 0.38 & 1.98/ \textbf{5.52}/ 0.04 & / & / & / \\ 
        IR & \textbf{11.66}/ 6.62/ \textbf{0.52} & 11.16/ 6.27/ 0.54 & 15.64/ 7.35/ 0.84 & 2.52/ 3.59/ 0.12 & / & / & / \\ 
        MI & 10.36/ 6.91/ 0.34 & 8.60/ 6.51/ 0.38 & 14.50/ 7.50/ 0.78 & 2.28/ 4.04/ 0.10 & 17.80/ 7.65/ 1.52 & 14.88/ 8.12/ 1.34 & 3.96/ 4.55/ 0.30 \\ 
        NI & 10.62/ \textbf{6.92}/ 0.48 & 8.66/ 6.53/ 0.46 & 14.12/ 7.49/ 0.74 & 2.38/ 4.01/ 0.10 & 18.14/ 7.71/ 1.80 & 14.54/ 8.11/ 1.32 & 4.12/ 4.57/ 0.28 \\ 
        VMI & 10.34/ 6.84/ 0.28 & 7.58/ 6.10/ 0.42 & 13.20/ 7.15/ 0.60 & \textbf{3.38}/ 0.06/ \textbf{1.40} & 9.40/ 5.87/ 0.60 & 7.54/ 5.92/ 0.54 & 3.86/ 4.47/ 0.32 \\ \hline
    \end{tabularx}  \vspace{-0.3cm}
\end{table}

\section{New Evaluation Protocols}
\vspace{-0.1cm}Since adversarial transferability is overestimated when tested only on CNNs~\cite{lecun1998gradient}, we propose a new benchmark including three protocols for reliable evaluation. 

\vspace{-0.1cm}
In \textbf{protocol-1}, an ensemble model is built based on 18 individual models where the output probabilities of the 18 models are averaged as the final output. The ensemble model is taken as a less-biased target model. The \textbf{protocol-2} counts the number of target models that AEs can transfer to, i.e., only the successfully attacked models are counted. A higher score of protocol-2 means higher transferability of AEs. The \textbf{protocol-3} considers a difficult setting where AE is considered as transferable if it can fool all 18 target models. 

\vspace{-0.1cm}
The results on the benchmark are shown in Tab.~\ref{protocol} where three scores in each cell correspond to protocol 1-3. PGD~\cite{madry2017towards} shows the lowest transferability in most cases, as expected. AE of TI~\cite{TI} attack shows low transferability, even lower than PGD when ViT-B/16~\cite{dosovitskiy2020image} and DeiT-B/16~\cite{touvron2021training} are taken as source models in protocol-1. When transferring from ViTs, DI~\cite{DIM} is the most effective method across all protocols. For most attack methods, AEs on ViTs show the highest transferability, namely, it is easier to transfer from ViT-B/16 and DeiT-B/16 to others. However, with ViTs as target models, the low transferability indicates that ViTs are more robust to AE created on others.

\vspace{-0.1cm}
Our protocols are comprehensive. For ILA~\cite{ILA} attack on IncV3, Tab.~\ref{protocol} shows that ILA can obtain the highest score on protocol-2, but it is the last but two when evaluated by protocol-1. Contrarily, AEs created by DI~\cite{DIM} attack on ViT-B/16 and Swin-B/4/7~\cite{liu2021swin} are more transferable than those generated on CNNs, and DI attack is the most efficient method for them. 

\section{Conclusion}
In this work, the adversarial transferability of AEs generated with representative attacks is re-evaluated across different types of DNNs, including CNNs, transformers, SNNs and DyNNs. The re-evaluation results are different from those evaluated only on CNNs from previous works and the transferability effectiveness of attack methods can be re-ranked. Based on our findings, we propose a new benchmark for a more reliable and comprehensive evaluation of adversarial transferability, which includes three protocols. The evaluation results on the new benchmark shows very low transferability of AEs, which indicates that adversarial transferability is overestimated before.

\vspace{-0.1cm}
\textbf{Limitation.} In this paper, we select 18 representative models and 12 representative transferable attacks from previous work of our community. Recently, more seminal models, e.g., capsule networks~\cite{gu2021effective} and vision-language models~\cite{gu2023towards} and more transferability-enhancing attack methods~\cite{Byun_2022_CVPR,gubri2022lgv,fang2022learning,zou2022making} have been proposed. Our current benchmark remains to be enriched. We release and maintain our benchmark to foster comprehensive evaluations of transferability by including more evaluation models and attack methods.

\vspace{-0.1cm}
\textbf{Social Impacts.} Our work re-evaluates the threats of transfer-based attacks. With our more comprehensive evaluation, the community can better understand the threats brought by DNNs.

{
\bibliographystyle{unsrt}
\bibliography{reference}
}

\end{document}